\documentclass[10pt,twocolumn,letterpaper]{article}

\usepackage{iccv}
\usepackage{times}
\usepackage{epsfig}
\usepackage{graphicx}
\usepackage{amsmath}
\usepackage{booktabs}
\usepackage{amssymb}
\usepackage{multirow}
\def\etal{\textit{et~al.}}

\usepackage[breaklinks=true,bookmarks=false]{hyperref}

\iccvfinalcopy 

\def\iccvPaperID{****} 

\ificcvfinal\fi

\begin{document}

\title{ Towards Large-scale Masked Face Recognition}

\author{
	 Manyuan Zhang$^{1,2}$,\hspace{0.3cm}   Bingqi Ma$^2$,\hspace{0.3cm} Guanglu Song$^2$,\hspace{0.3cm}  Yunxiao Wang$^2$, \hspace{0.3cm} Hongsheng Li$^1$, \hspace{0.3cm}  Yu Liu$^2$\\
	\\
	{$^1$Multimedia Laboratory, The Chinese University of Hong Kong} \\
	{$^2$BaseModel, SenseTime Group Limited} \hspace{0.3cm} \\
	{\texttt{\small{\ zhangmanyuan@link.cuhk.edu.hk}} \hspace{0.5cm}}
}

\maketitle
\ificcvfinal\thispagestyle{empty}\fi

\begin{abstract}

During the COVID-19 coronavirus epidemic, almost everyone is wearing masks, which poses a huge challenge for deep learning-based face recognition algorithms. In this paper, we will present our \textbf{championship} solutions in ICCV MFR WebFace260M and InsightFace unconstrained tracks. We will focus on four challenges in large-scale masked face recognition, i.e., super-large scale training,  data noise handling,  masked and non-masked face recognition accuracy balancing, and how to design inference-friendly model architecture. We hope that the discussion on these four aspects can guide future research towards more robust masked face recognition systems.

\end{abstract}

\section{Introduction}

Deep learning has deeply changed computer vision research~\cite{shi2023videoflow,deng2019arcface,wang2018cosface,shi2023flowformer++,zhang2023decoupled}.
The current state of face recognition is an exciting field of computer vision, with a diverse range of applications, such as access control and financial verification~\cite{deng2019arcface,liu2019towards,liu2021switchable,zhang2020discriminability,wang2018cosface,liu2017sphereface,zhang2022towards}. However, the COVID-19 pandemic has presented a significant challenge to the existing face recognition systems, as most individuals wear masks. Traditional face recognition algorithms tend to perform poorly under masked faces, highlighting the need for improved recognition techniques.

To foster the development of deep learning techniques, several competitions have been conducted~\cite{zhang2020top,chen20201st,caba2015activitynet,lin2014microsoft}. Deng et al.\cite{deng2021masked,zhu2021masked} organized the masked face recognition challenge at ICCV 2021, aimed at advancing masked face recognition in real-life scenarios. They introduced two novel benchmarks, the InsightFace track and WebFace260M track, for evaluating masked and unmasked face recognition. The benchmarks included real-world masked test sets, child test sets, and multi-ethnic test sets for comprehensive evaluation of masked face recognition. For more information about the benchmarks, please refer to the original paper\cite{deng2021masked,zhu2021masked}.

Previous work on masked face recognition has primarily focused on forming robust face recognition embeddings for masks. For example, Hariri~\cite{hariri2022efficient} extracted features only from non-masked regions, while Wang et al.\cite{wang2020hierarchical} and Li et al.\cite{li2021cropping} employed attention mechanisms to reduce the effect of masked regions on recognition features. However, these methods yielded limited performance in both the InsightFace track and WebFace260M track.

\begin{table*}[t]
\renewcommand\arraystretch{1.1} 
\begin{center}

\renewcommand\tabcolsep{1.2pt}
\begin{tabular}{cccccccc}
\toprule
Team                                                                                                & Mask   & Children & African & Caucasian & South Asian & East Asian & MR-ALL          \\ \hline
\begin{tabular}[c]{@{}c@{}}SenseTime Group Limited -\\ Base Model\end{tabular}                      & 94.731 & 93.123   & 98.366  & 99.092    & 98.679      & 92.322     & \textbf{98.487} \\
\begin{tabular}[c]{@{}c@{}}Alibaba DAMO Academy \&\\  National University of Singapore\end{tabular} & 94.163 & 94.412   & 98.373  & 99.010    & 99.014      & 91.638     & 98.322          \\
Toppan                                                                                              & 93.775 & 94.292   & 98.460  & 99.070    & 99.029      & 91.762     & 98.342          \\ 
\bottomrule
\end{tabular}
\caption{The final Leaderboard  for InsightFace unconstrained track. }
\label{tab:table1}
\end{center}
\end{table*}

\begin{table*}[t]
\renewcommand\arraystretch{1.1} 
\begin{center}

\renewcommand\tabcolsep{1.2pt}

\begin{tabular}{ccccccc}
\toprule
Team                                                                                          & ALL-Masked      & Wild-Masked & Controlled-Masked & ALL    & Wild   & Controlled      \\ \hline
\begin{tabular}[c]{@{}c@{}}SenseTime Group Limited -\\ Base Model\end{tabular}                & \textbf{0.0757} & 0.0959      & 0.0449            & 0.0148 & 0.0247 & 0.0017 \\
\begin{tabular}[c]{@{}c@{}}Face Group (David), AI Lab, \\ Kakao Enterprise\end{tabular}       & 0.0769          & 0.0990      & 0.0419            & 0.0241 & 0.0387 & 0.0021          \\
\begin{tabular}[c]{@{}c@{}}Alibaba DAMO Academy \&\\  NUS \& NTU \& AMS \& CASIA\end{tabular} & 0.0846          & 0.1036      & 0.0539            & 0.0145 & 0.0240 & 0.0016          \\ 
\bottomrule
\end{tabular}
\end{center}
\caption{The final Leaderboard  for WebFace260M track main result. }
\label{tab:table2}
\end{table*}

For masked face recognition, we employ mask augmentation~\cite{feng2018joint} on the training data to generate masked faces and add them to the training dataset to enhance the robustness of the algorithm for masked face recognition.
It is well known that the performance of a face recognition system is related to the scale of the training data~\cite{deng2019arcface,guo2016ms,cao2018celeb,cao2018vggface2,an2021partial}. In this challenge, we used the largest publicly available face dataset to date, WebFace42M, which is a cleaned subset of WebFace260M~\cite{zhu2021WebFace260M}. The WebFace42M contains 42M face photos from 2M identities, which is tens of times larger than traditional datasets such as MS1M~\cite{guo2016ms}. The super-large scale dataset poses a huge challenge to the face recognition system training. Therefore, how to deal with such large-scale face images and identities becomes the first challenge. WebFace42M is collected by crawlers from the Internet, and although it is semi-automatically cleaned, there is still a lot of noise in it, so how to deal with the noisy data is the second challenge. Increasing the proportion of masked augmentation in the training dataset will degrade the performance of non-masked face recognition, so how to trade off is the third challenge. In real industrial applications, inference time is very sensitive, so how to design inference-friendly model architecture and ensure efficient offline deployment is the fourth challenge.

After summarizing the four major challenges in achieving large-scale robust masked face recognition, we carefully design components to address these challenges. For the large-scale training challenge, we use distributed FC~\cite{deng2019arcface} and mixed precision~\cite{micikevicius2017mixed} for efficient training.  For the data noise, we use an iterative intra-class and inter-class cleaning strategy to make the data 
cleaner to train. As for the problem of masked and non-masked performance balance, we carefully adjust the masked augmentation ratio. Finally, for the inference time constraint, we use a latency-guided neural network search to search for the best model architecture within 1000ms of the CPU platform.

By successfully addressing the above challenges, \textbf{our models achieved championship results in both the InsightFace unconstrained and WebFace260M  tracks}, as shown in Table~\ref{tab:table1} and Table~\ref{tab:table2}. In the following, we will explore the details of our solution, which we hope will further inspire the face recognition community. 
\section{Method}

\subsection{Large scale training}

There is a consensus that as the training set for face recognition expands, the performance of the face recognition models also improves. From the early casia-webface~\cite{yi2014learning}, which contains 500,000 photos of 10,000 people, to the widely adopted MS1MV2~\cite{deng2019arcface,guo2016ms}, which contains 5.7 million photos of 85,000 identities, to WebFace42M~\cite{zhu2021WebFace260M}, which contains 42 million photos of 2 million identities, the proposed large-scale datasets have greatly contributed to the improvement of face recognition accuracy. Table~\ref{tab:table3} shows the results of the R100 model on different training sets, and it can be seen that the performance gradually improves as the size of the training set increases.

\begin{table}[t]
\renewcommand\arraystretch{1.1} 
\centering
\renewcommand\tabcolsep{1.3pt}
\begin{tabular}{cccccc}

\toprule
Data       & \# Id & \#Face & RFW   & Mega& IJB-C \\ \hline
MS1MV2~\cite{deng2019arcface}     & 85K   & 5.8M   & 98.98 & 98.40    & 96.03 \\
IMDB-Face~\cite{wang2018devil}  & 59K   & 1.7M   & 93.08 & 93.48    & 93.37 \\
WebFace42M~\cite{zhu2021masked} & 2M    & 42M    & \textbf{99.33} & \textbf{99.02}    & \textbf{97.70} \\ 

\bottomrule
\end{tabular}
\caption{Performance (\%) of different training data. R100 backbone
without flip test is adopted. RFW refers to average accuracy on~\cite{wang2019racial}, Mega refers to rank-1
identification on~\cite{kemelmacher2016megaface}, IJB-C is TAR@FAR=1e-4 on~\cite{maze2018iarpa}. }
\label{tab:table3}
\end{table}

\begin{table*}[t]
\renewcommand\arraystretch{1.1} 
\begin{center}

\renewcommand\tabcolsep{1.2pt}

\begin{tabular}{ccccc}
\toprule
Method            & Mixed-precision & All-Masked (MFR) & All (SFR)     & Training time (32*A100 GPU) \\ \hline
no Distributed FC &   $\times$   & out of memory    & out of memory & out of memory              \\
Distributed FC    &   $\times$   & -                & -             & 10 days                    \\
Distributed FC    &   $\checkmark$   & \textbf{0.0764 }           & \textbf{0.0146}        & 6 days                     \\
Partial FC        &   $\checkmark$   & 0.0786           & 0.0150        & \textbf{4 days}           \\
\bottomrule
\end{tabular}
\end{center}
\caption{The results of different training strategies on WebFace260M track.}
\label{tab:table4}
\end{table*}

\begin{figure*}[t]
    \centering
    \includegraphics[width=\linewidth]{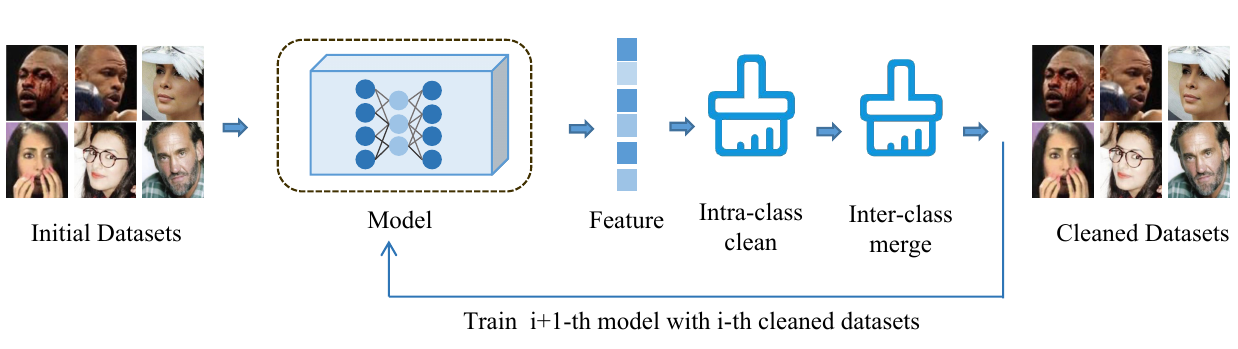}
    \caption{Our iterative data noise-cleaning process. We will first train a model $M_{i}$, then use it to extract features for the training set. Then do intra-class noise filtering and inter-class merging according to feature similarity. Then we will use the cleaned dataset to train model$M_{i+1}$ and repeat the cleaning process until performance converges.}
    \label{fig:fig1}
\end{figure*}

However, the large-scale dataset poses serious challenges for training. The first one is the huge pressure on the GPU memory due to the growth of the number of identities. For the optimization of the classification layer, whether using ArcFace~\cite{deng2019arcface} or CosFace~\cite{wang2018cosface} as the loss function, it is necessary to establish the softmax weight $W \in \mathbb{R}^{d \times C}$, where $d$ denotes embedding feature dimension and $C$ denotes the number of classes.  However, for millions of identities, the memory is not affordable for modern GPUs. So distributed FC was proposed~\cite{deng2019arcface}. A simple idea is to partition $W$ into $k$ sub-matrices $w$ of size $d \times \frac{C}{k}$ and place the $i$-th sub-matrices on the $i$-th GPU. To compute the overall softmax, we can first calculate the local sum of each GPU, and then compute the global by \textit{allgather}. And the size of communication overhead is equal to batch size multiplied by 4 bytes (Float32). 

The calculation of the GPU memory cost for  classification layer consists of two parts, $Mem_{w}$ and $M e m_{\text {logits }}$. The $Mem_{w}$ is calculated by $$
M e m_{w}=d \times \frac{C }{k } \times 4 \text { bytes } .
$$
where the $k$ is the number of GPU.

The storage of predicted logits on each GPU can be calculated as follows

$$M e m_{\text {logits }}=N k \times \frac{C}{k} \times 4   \ \text{bytes}$$ 

where $N$ is the mini-batch size on each GPU.

For the optimizer SGD, each parameter occupies 12 bytes; for CosFace~\cite{wang2018cosface} and ArcFace~\cite{deng2019arcface}, each logit occupies 8 bytes. so the GPU memory occupied by a single classification layer is

$$\mathrm{Mem}_{F C}=3 \times M e m_{W}+2 \times M e m_{\text {logits }}$$

For the WebFace42M dataset, if we don't use distributed FC, the GPU memory usage of a single classification layer will be 12.44 GB (2 million IDs, 512 dimensions,  batch size 64 per GPU), which is unacceptable. For distributed FC, if 32 GPU cards are used for training, the memory consumption of the classification layer is only 1.3 GB, which greatly reduces the burden. 

An~\etal~\cite{an2021partial} further introduces the partial FC, which randomly samples the negative class samples so introduces much speed up, and reduces memory consumption. 

Besides the high GPU memory cost for the classification layer, another challenge is the training cost, since WebFace42M  contains 42M face images, a full training with  A100 will take 10 days. By applying the mixed-precision training~\cite{micikevicius2017mixed}, which means utilizing FP16 to backbone parameters and FP32 to classification parameters, the training speed can speed up to above 30\%. The results comparison for Distributed FC and Partial FC and mixed training are shown in Table~\ref{tab:table4}.

\subsection{Iterative data noise cleaning}
\label{iter_data}

\begin{figure}[t]
    \centering
    \includegraphics[width=\linewidth]{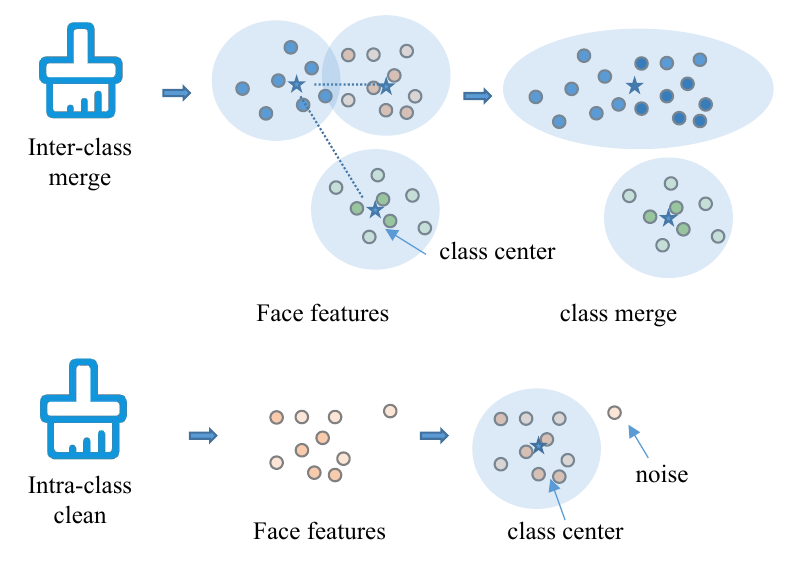}
    \caption{The process of intra-class filtering and inter-class merging. We filter face images whose feature similarity with class center is smaller than the threshold $thre\_intra$. Then we merge those class pairs whose class center feature similarity is greater than the threshold $thre\_inter$. The threshold $thre\_intra$ and threshold $thre\_inter$ are pre-defined and can be searched according to clean results. The class center feature is simply calculated by the average pooling of features of class images. }
    \label{fig:fig2}
\end{figure}

Although WebFace42M is self-cleaned, there is still a lot of noise in it. These noises will harm training. The noise of face training is divided into two categories~\cite{wang2018devil}, namely the `outliers' and `label flips'. An `outlier' noise means an image does not belong to any class of the datasets.
A `label flip' noise refers to an image that is wrongly labeled with the incorrect class label. For these two types of noise, we design a self-iterating cleanup strategy based on face features. That is, we first train model $M_{i}$ with the current data, then use model  $M_{i}$ to extract features for the training set, and then filter outliers based on the similarity between the image feature and its class center feature (obtained by averaging all sample features of the class), with a filtering threshold of $thre\_intra$. After intra-class filtering, we update the class centers with the remaining samples and perform a similarity 
search on the class centers,  then merge the class center pairs with a similarity threshold 
higher than $thre\_inter$. We retrain the model using the cleaned data to get model $M_{i+1}$, and then repeat the process until there is no performance gain. This iterative process is shown in Figure~\ref{fig:fig1}. The process of the inter-class clean and inter-class merge is shown in Figure~\ref{fig:fig2}.

Because this iterative process will consume a lot of computing, we use the  R100 as the backbone. We performed two iterations on WebFace42M, and the results are shown in Table~\ref{tab:table5}. This cleaning process will gain more on dirtier datasets such as MS1M~\cite{guo2016ms}. 

\subsection{Mask augmentation and ratio trade-off}

\begin{table}[t]
\begin{center}

\begin{tabular}{cc}
\toprule
         & \begin{tabular}[c]{@{}l@{}}Results on \\ All-Masked(MFR)\end{tabular} \\ \hline
un-clean & 0.1774                                                                \\
clean    & \textbf{0.1735 }                   \\
\bottomrule
\end{tabular}
\caption{The results of iterative cleaning on the MFR WebFace260M track. The backbone used for iteration is R100.}
\label{tab:table5}
\end{center}
\end{table}

\begin{figure}
    \centering
    \includegraphics[width=\linewidth]{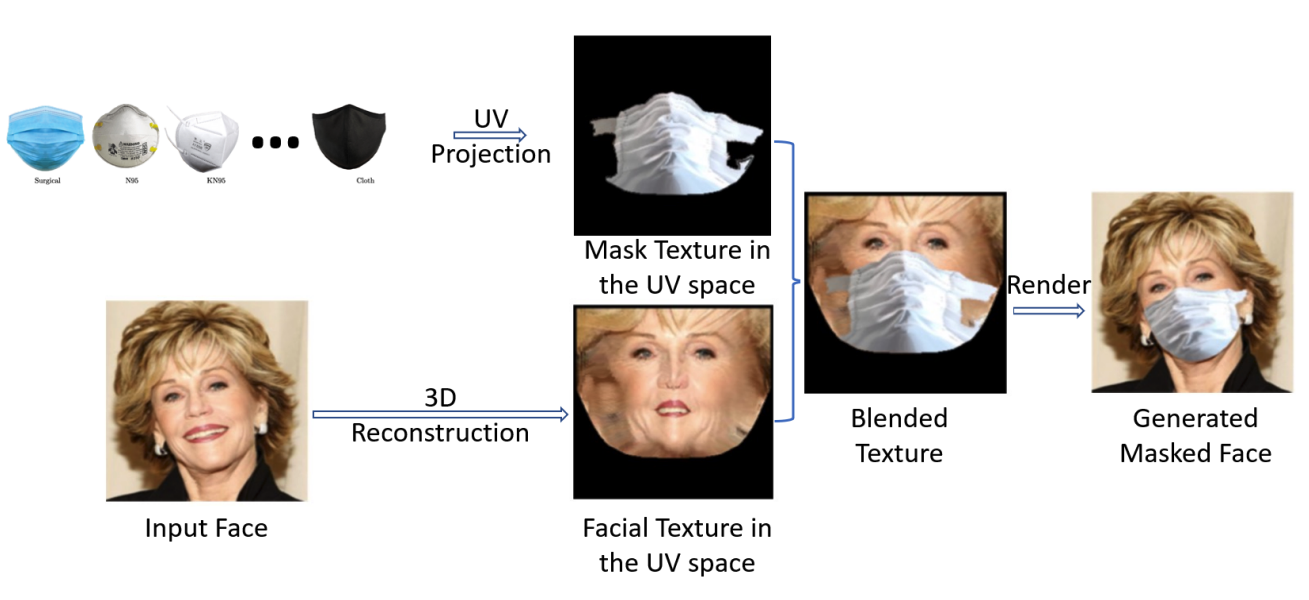}
    \caption{Masked face augmentation pipeline. Demo images and the tool are from ~\cite{deng2019arcface,wang2021facex}}.
    \label{fig:fig3}
\end{figure}

To achieve a more robust performance on masked face recognition, we need to perform mask augmentation on the training set images. We follow the  JDAI-CV toolkit~\cite{wang2021facex} provided by Wang~\etal. which is based on PRNet~\cite{feng2018joint} for mask addition. The process of mask addition is shown in Figure~\ref{fig:fig3}. First, we do a 3D reconstruction based on the 2D face image. We will get the UV texture map, the face geometry, and the camera pose. Then we pick one of the mask templates at random and project it into the UV space. Based on texture blending, we can easily get the masked facial UV texture, and finally, we combine the masked facial UV texture and the face geometry to render the face image in 2D.

\begin{figure}
    \centering
    \includegraphics[width=0.8\linewidth]{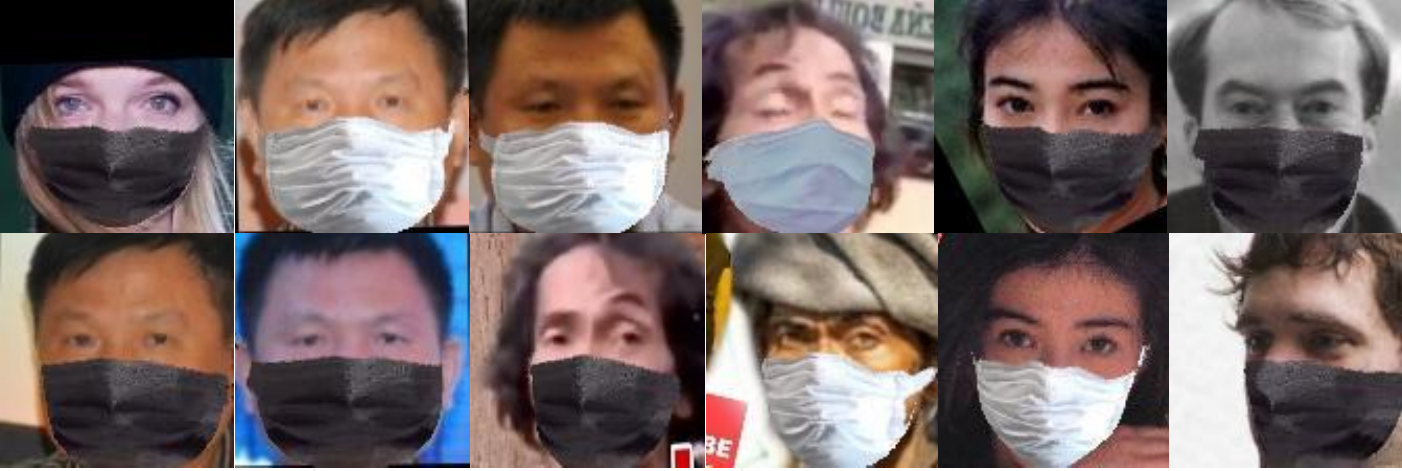}
    \caption{Some examples of mask augmentation via JDAI-CV toolkit~\cite{wang2021facex}. }
    \label{fig:fig4}
\end{figure}

Some examples of mask augmentation are shown in Figure~\ref{fig:fig4}. As we can see from the figure, the addition of the mask is very realistic.

However, mask augmentation is in conflict with non-masked face recognition, and mask augmentation of too many images in the training set will impair non-masked face recognition, so how to balance the ratio of additions is a trade-off. We tried different ratios, and the results are shown in Table~\ref{tab:tab4}. 15\% of the augmentations achieved the most balanced results, so we added masks to 15\% of the images in all training sets, with a random sampling of the mask addition styles.

\begin{table}[]

\renewcommand\arraystretch{1.1} 
\centering
\begin{tabular}{ccc}
\toprule

Mask Ratio & All-Maksed (MFR) & All (SFR) \\ \hline
10\%       & 0.0872           & \textbf{0.0163}    \\
15\%       & \textbf{0.0860  }         & 0.0183    \\
20\%       & 0.0931           & 0.0311   \\
\bottomrule
\end{tabular}
\caption{The validation of different mask augmentation ratios. The results are reported on WebFace260M track, 15\% achieved the best performance.}
\label{tab:tab4}
\end{table}

\subsection{Latency guided neural architecture search }

The speed of inferencing is also very sensitive in industrial applications. In access control or mobile authentication scenarios, real-time is an important guarantee for the application. Therefore, designing inference-friendly model architecture is very important. In this competition, the InsightFace unconstrained track limits the GPU inference time to 30ms (V100) and the model size to 1GB, while the WebFace260M track limits the CPU inference time to 1000ms for a single thread (Intel Xeon CPU E5-2630-v4@2.20GHz processor). In order to be able to design latency time-friendly architecture, we have conducted a neural architecture search (NAS). MNasNet~\cite{tan2019mnasnet} uses reinforcement learning to guide the network search process and use latency on the target platform as a reward. We follow it.

We consider the accuracy of the proxy set and the latency on the target platform. To this end, we use a weighted product method to approximate Pareto optimal solutions. The reward function is formulated as :

\begin{equation} \label{eq_R}
    \texttt{R}(m) = {{\texttt{ACC}(m)} \times {\left[{\frac{\texttt{COST}(m)}{\texttt{TAR}}}\right]^\alpha}} ,
\end{equation}
where $\texttt{ACC}(m)$ is the accuracy on the proxy task, $\texttt{COST}(m)$ is the computation cost (FLOPs) of the architecture, $\texttt{TAR}$ is the target computation cost, and $\alpha$ is the weight factor.  

In order to design a more latency-friendly backbone in CPU (most deployment scenarios), we carefully design the search space, which is split-attention based.

We summarize the path architecture as follows:
\begin{itemize}
    \item \textit{Kernel size},  regular kernel size containing 1x1, 3x3, 5x5, and 7x7. Asymmetric kernel size containing 1x3+3x1, 1x5+5x1, and 1x7+7x1
    \item \textit{Path depth}, depth 2, 3, and 4 are covered. 
    \item \textit{Densely connected}, whether the convolution layers of the path are densely connected.
\end{itemize}

It is worth mentioning that we \textbf{do not} add \textit{depth-wise convolution} in our search space due to its huge GPU memory consumption and inefficient inference latency. Finally, we get our RIDNet-S, and we test it in InsightFace track IJB-B and IJB-C, the results are shown in  Table~\ref{tab:table7}. For different latency constraints, we upscale the width, depth, and input resolution~\cite{tan2019efficientnet}. Target at 1000 ms (Intel Xeon CPU E5-2630-v4@2.20GHz processor) we get RIDNet-M which is of 30.85G FLOPS and 161M params and finally achieve champion results on InsightFace unconstrained and WebFace 260M track shown in Table~\ref{tab:table1} and Table~\ref{tab:table2}. By continuing to scale up the model, we get the RIDNet-L, which is 71.58G FLOPS and 282M params. It achieves the state-of-the-art performance of the MFR-ongoing leaderboard (ongoing version of InsightFace track), and surpasses the second place (team `pnan`, submitted on 2022-04-01) hugely, as shown in Table~\ref{tab:table8}.

\begin{table*}[]
\renewcommand\arraystretch{1.1} 
\centering
\begin{tabular}{ccccccccc}
\toprule

\multirow{2}{*}{Model} & \multirow{2}{*}{FLOPS} & \multirow{2}{*}{Params} & \multirow{2}{*}{\begin{tabular}[c]{@{}c@{}}IJB-B\\ 1e-4\end{tabular}} & \multirow{2}{*}{\begin{tabular}[c]{@{}c@{}}IJB-C\\ 1e-4\end{tabular}} & \multicolumn{4}{c}{MFR On-going}                      \\ 
                       &                        &                         &                                                                       &                                                                       & African & Caucasian & East Asian & SouthAsian \\ \hline
ResNet-50              & 4.12G                  & 23.5M                   & 93.76                                                                 & 95.21                                                                 & 72.08   & 82.48     & 80.6       & 53.27      \\
SE-ResNet-50           & 4.12G                  & 26.0M                   & 94.3                                                                  & 95.64                                                                 & 74.51   & 85.49     & 81.74      & 56.02      \\
ResNeXt-50             & 4.27G                  & 22.9M                   & 93.99                                                                 & 95.53                                                                 & 73.26   & 83.63     & 77.42      & 56.1       \\
R50                    & 6.32G                  & 25.4M                   & 94.35                                                                 & 95.67                                                                 & 74.22   & 84.35     & 82.18      & 56.43      \\
RIDNet-S                 & 5.34G                  & \textbf{19.7M }                  & \textbf{95.43}                                                                 & \textbf{96.72}                                                                 & \textbf{81.98}   & \textbf{88.38 }    & \textbf{87.59 }     & \textbf{61.02}     \\
\bottomrule
\end{tabular}
\caption{Comparison of different architecture with our RIDNet-S trained on MS1M. Accuracy, FLOPS, and params are reported.}
\label{tab:table7}
\end{table*}

\begin{table*}[t]
\renewcommand\arraystretch{1.1} 
\begin{center}

\renewcommand\tabcolsep{1.2pt}
\begin{tabular}{cccccccc}
\toprule
Team                                                                                                & Mask   & Children & African & Caucasian & South Asian & East Asian & MR-ALL          \\ \hline
\begin{tabular}[c]{@{}c@{}}RIDNet-M\\ Submitted to ICCV MFR 2021\end{tabular}                      & 94.731 & 93.123   & 98.366  & 99.092    & 98.679      & 92.322     & 98.487 \\
\begin{tabular}[c]{@{}c@{}}RIDNet-L \\  \end{tabular} & \textbf{95.972} & 94.042  & \textbf{98.630}  & \textbf{99.233}    & \textbf{99.196}     & \textbf{93.339 }    &\textbf{ 98.691 }         \\
\begin{tabular}[c]{@{}c@{}}pnan \(\text{2022-04-01}\) \\  \end{tabular}                                                                                                &  94.464 & \textbf{94.237}   & 98.494   &   99.105  &  98.979     &  92.512    &   98.467        \\ 
\bottomrule
\end{tabular}
\caption{The  Leaderboard for MFR-Ongoing. By upscaling our RIDNet, we achieve state-of-the-art results. }
\label{tab:table8}
\end{center}
\end{table*}

\begin{table}[h]
\renewcommand\arraystretch{1.1} 
\centering
\renewcommand\tabcolsep{1.2pt}
\begin{tabular}{cccc}
\toprule
                                                    &                & All-Maksed (MFR) & All (SFR) \\ \hline
\multicolumn{1}{c|}{\multirow{3}{*}{Loss function}} & CosFace m=0.4  & \textbf{0.0873  }         & \textbf{0.0168  }  \\
\multicolumn{1}{c|}{}                               & ArcFace m=0.5  & 0.0979           & 0.0196    \\
\multicolumn{1}{c|}{}                               & ArcFace m=0.45 & 0.1005           & 0.0201    \\ \hline
\multicolumn{1}{c|}{\multirow{2}{*}{Decay}}         & 1e-4           & \textbf{0.1340   }        & \textbf{0.0242}    \\
\multicolumn{1}{c|}{}                               & 5e-4           & 0.1585           & 0.0488    \\ \hline
\multicolumn{1}{c|}{\multirow{2}{*}{Dropout}}       & Ratio 0.4      & \textbf{0.0786   }        & \textbf{0.0150 }   \\
\multicolumn{1}{c|}{}                               & no dropout     & 0.0792           & 0.0158   \\ \bottomrule
\end{tabular}
\caption{The results of hyper-parammeter tuning on WebFace260M track.}
\label{tab:table9}
\end{table}

\begin{table}[h]
\renewcommand\arraystretch{1.1} 
\centering
\renewcommand\tabcolsep{1.2pt}
\begin{tabular}{ccc}
\toprule
Finetune          & All-Masked (MFR) & All (SFR)       \\ \hline
Before  & 0.0764           & \textbf{0.0145} \\
After   & \textbf{0.0757}  & 0.0148         \\
\bottomrule
\end{tabular}
\caption{Results of finetuning on WebFace260M track. To further improve the performance of masked face recognition, we finetune the model with mask augmentation 33\%.}
\label{tab:table10}
\end{table}

=

\section{Experiment details}

In this section, we will introduce the experiment details of our solutions. including the training dataset WebFace42M, the test protocol of two tracks. We will also show our training setting, as well as the ablation study of hyper-parameters. We also conduct fine-tuning,  which further improves the performance of the masked face.

\subsection{Training dataset}

The training set we used is WebFace42M, which is the subset of the largest public face recognition datasets WebFace260M~\cite{zhu2021WebFace260M}. It is self-cleaned and aims to close the data gap between academia and industry. It first collects the celebrity name list, which consists of two-part, one is MS1M and the other is IMDB. Based on these two lists, they search the celebrity faces via Google images and get the WebFace260M. Then the WebFace42M was obtained by the CAST pipeline.  The WebFace42M is now the largest noise-control recognition dataset, however as we have claimed in section~\ref{iter_data}, to conduct our carefully designed iterative data noise cleaning, the performance can be further improved.

\subsection{Evaluation protocol}

\paragraph{InsightFace Track}

There are three test sets, the masked test set, the children test set, and the multi-racial test Set. The statistics of the test sets can be viewed in the original paper~\cite{deng2021masked}. Note that the images have few overlaps with the popular face training set, which guarantees fairness. All challenge submissions are ordered in terms of weighted TPRs across two test sets (i.e. Masked Test Set and Multi-racial Test Set) by the formula of 0.25 * TPR@Masked + 0.75 * TPR@MR-All.  After the competition, the InsightFace Track has been re-opened as the MFR-on-going test, which is a large-scale and authoritative face recognition benchmark.
For the unconstrained track, there is no training data limitation. We train our model on WebFace42M.

\paragraph{WebFace260M Track}

The identities pool of WebFace260M Track consists of 2,748 celebrities. There are three types of test images, the controlled image such as visa or driving license, the wild image which is collected in unconstrained scenarios, and the cross-age image, which covers a huge age gap. There are elaborately constructed 2,478 identities and 57,715 faces
in total for SFR. 3,328,950,920 impostors and 2,070,305
genuine pairs could be constructed. As for MFR, 3,211 masked faces celebrities among 862 identities are selected. The final  competition ranking is All
(MFR\&SFR) metric: All (MFR\&SFR)= 0.25 $\times$ All-Masked + 0.75 $\times$ All (SFR). Note that this track constrains the training set, participants must only use WebFace260M.

\subsection{Training setting}

In the experiments above, the most common training configuration we use is listed as follows. We use the SGD with momentum 0.9 as our optimizer. The base learning rate is 0.1 for 256 and linear scale-up. The learning rate decays at 5, 9, 12, and 15 epochs, and the training will end at 17 epochs. We use a total batch size of 4512 (48 NVIDIA A100 ) and the embedding dim is 512.

\subsection{Hyper-parameter tuning}

We do some hyper-parameter tuning and the results are shown in  Table~\ref{tab:table9}. As we can see from the table, CosFace~\cite{wang2018cosface} with a margin of 0.4 achieves the best performance. The proper weight decay is 1e-4 and the results are very sensitive to the weight decay. As for the dropout, 0.4 achieves higher performance.

\subsection{Finetune}

To further improve the performance on masked face recognition, we finetune our model trained with mask augmentation ratio 15\% with mask augmentation ratio 33\% for 3 epochs. As shown in  Table~\ref{tab:table10}, even the SFR drops, however, the combination of ALL and Masked has been further improved. This is our final submission version.

\section{Conclusion}

During the COVID-19 coronavirus, the demand for robust mask face recognition algorithms is gradually rising. In this paper, we enumerate four challenges for implementing large-scale masked face recognition, namely, training difficulties due to super-large training sets, training set pollution due to data noise, robust mask enhancement algorithms, and deployment inference time constraints. We propose methods to address each of these challenges. We hope that these insights can further advance the face recognition community.


{\small
\bibliographystyle{ieee_fullname}
\bibliography{main}

\begin{thebibliography}{10}\itemsep=-1pt

\bibitem{an2021partial}
Xiang An, Xuhan Zhu, Yuan Gao, Yang Xiao, Yongle Zhao, Ziyong Feng, Lan Wu, Bin
  Qin, Ming Zhang, Debing Zhang, et~al.
\newblock Partial fc: Training 10 million identities on a single machine.
\newblock In {\em Proceedings of the IEEE/CVF International Conference on
  Computer Vision}, pages 1445--1449, 2021.

\bibitem{caba2015activitynet}
Fabian Caba~Heilbron, Victor Escorcia, Bernard Ghanem, and Juan Carlos~Niebles.
\newblock Activitynet: A large-scale video benchmark for human activity
  understanding.
\newblock In {\em Proceedings of the ieee conference on computer vision and
  pattern recognition}, pages 961--970, 2015.

\bibitem{cao2018celeb}
Jiajiong Cao, Yingming Li, and Zhongfei Zhang.
\newblock Celeb-500k: A large training dataset for face recognition.
\newblock In {\em 2018 25th IEEE International Conference on Image Processing
  (ICIP)}, pages 2406--2410. IEEE, 2018.

\bibitem{cao2018vggface2}
Qiong Cao, Li Shen, Weidi Xie, Omkar~M Parkhi, and Andrew Zisserman.
\newblock Vggface2: A dataset for recognising faces across pose and age.
\newblock In {\em 2018 13th IEEE international conference on automatic face \&
  gesture recognition (FG 2018)}, pages 67--74. IEEE, 2018.

\bibitem{chen20201st}
Siyu Chen, Junting Pan, Guanglu Song, Manyuan Zhang, Hao Shao, Ziyi Lin, Jing
  Shao, Hongsheng Li, and Yu Liu.
\newblock 1st place solution for ava-kinetics crossover in acitivitynet
  challenge 2020.
\newblock {\em arXiv preprint arXiv:2006.09116}, 2020.

\bibitem{deng2021masked}
Jiankang Deng, Jia Guo, Xiang An, Zheng Zhu, and Stefanos Zafeiriou.
\newblock Masked face recognition challenge: The insightface track report.
\newblock In {\em Proceedings of the IEEE/CVF International Conference on
  Computer Vision}, pages 1437--1444, 2021.

\bibitem{deng2019arcface}
Jiankang Deng, Jia Guo, Niannan Xue, and Stefanos Zafeiriou.
\newblock Arcface: Additive angular margin loss for deep face recognition.
\newblock In {\em Proceedings of the IEEE/CVF conference on computer vision and
  pattern recognition}, pages 4690--4699, 2019.

\bibitem{feng2018joint}
Yao Feng, Fan Wu, Xiaohu Shao, Yanfeng Wang, and Xi Zhou.
\newblock Joint 3d face reconstruction and dense alignment with position map
  regression network.
\newblock In {\em Proceedings of the European conference on computer vision
  (ECCV)}, pages 534--551, 2018.

\bibitem{guo2016ms}
Yandong Guo, Lei Zhang, Yuxiao Hu, Xiaodong He, and Jianfeng Gao.
\newblock Ms-celeb-1m: A dataset and benchmark for large-scale face
  recognition.
\newblock In {\em European conference on computer vision}, pages 87--102.
  Springer, 2016.

\bibitem{hariri2022efficient}
Walid Hariri.
\newblock Efficient masked face recognition method during the covid-19
  pandemic.
\newblock {\em Signal, image and video processing}, 16(3):605--612, 2022.

\bibitem{kemelmacher2016megaface}
Ira Kemelmacher-Shlizerman, Steven~M Seitz, Daniel Miller, and Evan Brossard.
\newblock The megaface benchmark: 1 million faces for recognition at scale.
\newblock In {\em Proceedings of the IEEE conference on computer vision and
  pattern recognition}, pages 4873--4882, 2016.

\bibitem{li2021cropping}
Yande Li, Kun Guo, Yonggang Lu, and Li Liu.
\newblock Cropping and attention based approach for masked face recognition.
\newblock {\em Applied Intelligence}, 51(5):3012--3025, 2021.

\bibitem{lin2014microsoft}
Tsung-Yi Lin, Michael Maire, Serge Belongie, James Hays, Pietro Perona, Deva
  Ramanan, Piotr Doll{\'a}r, and C~Lawrence Zitnick.
\newblock Microsoft coco: Common objects in context.
\newblock In {\em European conference on computer vision}, pages 740--755.
  Springer, 2014.

\bibitem{liu2021switchable}
Boxiao Liu, Guanglu Song, Manyuan Zhang, Haihang You, and Yu Liu.
\newblock Switchable k-class hyperplanes for noise-robust representation
  learning.
\newblock In {\em Proceedings of the IEEE/CVF International Conference on
  Computer Vision}, pages 3019--3028, 2021.

\bibitem{liu2017sphereface}
Weiyang Liu, Yandong Wen, Zhiding Yu, Ming Li, Bhiksha Raj, and Le Song.
\newblock Sphereface: Deep hypersphere embedding for face recognition.
\newblock In {\em Proceedings of the IEEE conference on computer vision and
  pattern recognition}, pages 212--220, 2017.

\bibitem{liu2019towards}
Yu Liu et~al.
\newblock Towards flops-constrained face recognition.
\newblock In {\em Proceedings of the IEEE/CVF International Conference on
  Computer Vision Workshops}, pages 0--0, 2019.

\bibitem{maze2018iarpa}
Brianna Maze, Jocelyn Adams, James~A Duncan, Nathan Kalka, Tim Miller, Charles
  Otto, Anil~K Jain, W~Tyler Niggel, Janet Anderson, Jordan Cheney, et~al.
\newblock Iarpa janus benchmark-c: Face dataset and protocol.
\newblock In {\em 2018 international conference on biometrics (ICB)}, pages
  158--165. IEEE, 2018.

\bibitem{micikevicius2017mixed}
Paulius Micikevicius, Sharan Narang, Jonah Alben, Gregory Diamos, Erich Elsen,
  David Garcia, Boris Ginsburg, Michael Houston, Oleksii Kuchaiev, Ganesh
  Venkatesh, et~al.
\newblock Mixed precision training.
\newblock {\em arXiv preprint arXiv:1710.03740}, 2017.

\bibitem{shi2023videoflow}
Xiaoyu Shi, Zhaoyang Huang, Weikang Bian, Dasong Li, Manyuan Zhang, Ka~Chun
  Cheung, Simon See, Hongwei Qin, Jifeng Dai, and Hongsheng Li.
\newblock Videoflow: Exploiting temporal cues for multi-frame optical flow
  estimation.
\newblock {\em arXiv preprint arXiv:2303.08340}, 2023.

\bibitem{shi2023flowformer++}
Xiaoyu Shi, Zhaoyang Huang, Dasong Li, Manyuan Zhang, Ka~Chun Cheung, Simon
  See, Hongwei Qin, Jifeng Dai, and Hongsheng Li.
\newblock Flowformer++: Masked cost volume autoencoding for pretraining optical
  flow estimation.
\newblock In {\em Proceedings of the IEEE/CVF Conference on Computer Vision and
  Pattern Recognition}, pages 1599--1610, 2023.

\bibitem{tan2019mnasnet}
Mingxing Tan, Bo Chen, Ruoming Pang, Vijay Vasudevan, Mark Sandler, Andrew
  Howard, and Quoc~V Le.
\newblock Mnasnet: Platform-aware neural architecture search for mobile.
\newblock In {\em Proceedings of the IEEE/CVF Conference on Computer Vision and
  Pattern Recognition}, pages 2820--2828, 2019.

\bibitem{tan2019efficientnet}
Mingxing Tan and Quoc Le.
\newblock Efficientnet: Rethinking model scaling for convolutional neural
  networks.
\newblock In {\em International conference on machine learning}, pages
  6105--6114. PMLR, 2019.

\bibitem{wang2018devil}
Fei Wang, Liren Chen, Cheng Li, Shiyao Huang, Yanjie Chen, Chen Qian, and
  Chen~Change Loy.
\newblock The devil of face recognition is in the noise.
\newblock In {\em Proceedings of the European Conference on Computer Vision
  (ECCV)}, pages 765--780, 2018.

\bibitem{wang2018cosface}
Hao Wang, Yitong Wang, Zheng Zhou, Xing Ji, Dihong Gong, Jingchao Zhou, Zhifeng
  Li, and Wei Liu.
\newblock Cosface: Large margin cosine loss for deep face recognition.
\newblock In {\em Proceedings of the IEEE conference on computer vision and
  pattern recognition}, pages 5265--5274, 2018.

\bibitem{wang2021facex}
Jun Wang, Yinglu Liu, Yibo Hu, Hailin Shi, and Tao Mei.
\newblock Facex-zoo: A pytorch toolbox for face recognition.
\newblock In {\em Proceedings of the 29th ACM International Conference on
  Multimedia}, pages 3779--3782, 2021.

\bibitem{wang2019racial}
Mei Wang, Weihong Deng, Jiani Hu, Xunqiang Tao, and Yaohai Huang.
\newblock Racial faces in the wild: Reducing racial bias by information
  maximization adaptation network.
\newblock In {\em Proceedings of the ieee/cvf international conference on
  computer vision}, pages 692--702, 2019.

\bibitem{wang2020hierarchical}
Qiangchang Wang, Tianyi Wu, He Zheng, and Guodong Guo.
\newblock Hierarchical pyramid diverse attention networks for face recognition.
\newblock In {\em Proceedings of the IEEE/CVF conference on computer vision and
  pattern recognition}, pages 8326--8335, 2020.

\bibitem{yi2014learning}
Dong Yi, Zhen Lei, Shengcai Liao, and Stan~Z Li.
\newblock Learning face representation from scratch.
\newblock {\em arXiv preprint arXiv:1411.7923}, 2014.

\bibitem{zhang2020top}
Manyuan Zhang, Hao Shao, Guanglu Song, Yu Liu, and Junjie Yan.
\newblock Top-1 solution of multi-moments in time challenge 2019.
\newblock {\em arXiv preprint arXiv:2003.05837}, 2020.

\bibitem{zhang2022towards}
Manyuan Zhang, Guanglu Song, Yu Liu, and Hongsheng Li.
\newblock Towards robust face recognition with comprehensive search.
\newblock In {\em European Conference on Computer Vision}, pages 720--736.
  Springer, 2022.

\bibitem{zhang2023decoupled}
Manyuan Zhang, Guanglu Song, Yu Liu, and Hongsheng Li.
\newblock Decoupled detr: Spatially disentangling localization and
  classification for improved end-to-end object detection.
\newblock In {\em Proceedings of the IEEE/CVF International Conference on
  Computer Vision}, pages 6601--6610, 2023.

\bibitem{zhang2020discriminability}
Manyuan Zhang, Guanglu Song, Hang Zhou, and Yu Liu.
\newblock Discriminability distillation in group representation learning.
\newblock In {\em European Conference on Computer Vision}, pages 1--19.
  Springer, 2020.

\bibitem{zhu2021masked}
Zheng Zhu, Guan Huang, Jiankang Deng, Yun Ye, Junjie Huang, Xinze Chen, Jiagang
  Zhu, Tian Yang, Jia Guo, Jiwen Lu, et~al.
\newblock Masked face recognition challenge: The webface260m track report.
\newblock {\em arXiv preprint arXiv:2108.07189}, 2021.

\bibitem{zhu2021WebFace260M}
Zheng Zhu, Guan Huang, Jiankang Deng, Yun Ye, Junjie Huang, Xinze Chen, Jiagang
  Zhu, Tian Yang, Jiwen Lu, Dalong Du, et~al.
\newblock Webface260m: A benchmark unveiling the power of million-scale deep
  face recognition.
\newblock In {\em Proceedings of the IEEE/CVF Conference on Computer Vision and
  Pattern Recognition}, pages 10492--10502, 2021.

\end{thebibliography}
}

\end{document}